\documentclass[conference]{IEEEtran}
\IEEEoverridecommandlockouts

\usepackage{subfigure}
\usepackage{sidecap}
\usepackage{wrapfig}

\usepackage{multirow}
\usepackage{caption}

\usepackage{cite}
\usepackage{amsmath,amssymb,amsfonts}
\usepackage{algorithmic}
\usepackage{graphicx}
\usepackage{textcomp}
\usepackage{xcolor}

\usepackage{subfigure}
\usepackage{sidecap}
\usepackage{wrapfig}
\usepackage{multirow}

\begin{document}

\title{Feature Relevance Analysis to Explain Concept Drift - A Case Study in Human Activity Recognition}

\author{\IEEEauthorblockN{Pekka Siirtola}
\IEEEauthorblockA{\textit{Biomimetics and Intelligent Systems Group} \\
\textit{University of Oulu}\\
Oulu, Finland \\
pekka.siirtola@oulu.fi}
\and
\IEEEauthorblockN{Juha Röning}
\IEEEauthorblockA{\textit{Biomimetics and Intelligent Systems Group} \\
\textit{University of Oulu}\\
Oulu, Finland \\
juha.roning@oulu.fi}
}

\maketitle

\begin{abstract}
This article studies how to detect and explain concept drift. Human activity recognition is used as a case study together with a online batch learning situation where the quality of the labels used in the model updating process starts to decrease. Drift detection is based on identifying a set of features having the largest relevance difference between the drifting model and a model that is known to be accurate and monitoring how the relevance of these features changes over time. As a main result of this article, it is shown that feature relevance analysis cannot only be used to detect the concept drift but also to explain the reason for the drift when a limited number of typical reasons for the concept drift are predefined. To explain the reason for the concept drift, it is studied how these predefined reasons effect to feature relevance. In fact, it is shown that each of these has an unique effect to features relevance and these can be used to explain the reason for concept drift.
\end{abstract}


\begin{IEEEkeywords}
Human activity recognition, personalizing, incremental learning, online learning, feature importance analysis, accelerometer.
\end{IEEEkeywords}


\section{Problem statement and related work}
\label{problem}

This study focuses on human activity recognition based on inertial sensor data collected using smartphone sensors. There are already a lot of wellness wearables in the market which rely on sensor data. Most of them are using static pre-build algorithms, and thus, they are not able to adapt to unseen situations. In fact, the most common way to build a machine learning prediction model is to rely on data that are given prior to training the model. The problem of this approach is that it assumes that structure of the data remains static. However, this is not the case in the real-world problems as the world around us constantly changes. Moreover, while pre-build algorithms provide high recognition rates on most of the people, but not for all \cite{albert2012using}. Due to this, the recognition should be based on adaptive personalized models and not on static user-independent models.

In the case of wellness applications, incremental learning can be used to personalize and improve recognition models, and adapt them to new environments. When it comes to ensemble-based incremental learning models, like the ones studied in this article, personalizing by learning new personal base models from streaming data, and adding these to ensemble \cite{hammerchoosing}. Therefore, in the case of incremental learning, the recognition model is updated, and model re-training is not needed. In fact, it has been shown that wearable sensor data based human activity recognition models benefit from incremental learning as there are some studies where incremental learning is used to recognize human activities based on wearable sensor data (\cite{Mo2016}, \cite{Ntalampiras} and \cite{wang2012incremental}). These articles do not focus on personalizing human activity recognition models, but it has shown in related studies that incremental learning can be used for that as well.

In \cite{mazankiewicz2020incremental} and \cite{siirtola2018ESANN}  unsupervised methods to personalize human activity recognition models without user-interruption based on incremental learning were presented. For instance, \cite{siirtola2018ESANN} is based on ensemble method which processes the incoming streaming data as batches. For each batch, a new group of weak base models are trained, and combined to a group of previously trained models. In the context of human activity recognition, the problem is that true labels are not available. Therefore, when it comes to incremental learning, there are two options for obtaining labels for the online data: ask labels from the user, or train new models based on predicted labels. However, considering predicted labels as true labels is risky as it is not sure if they are correctly classified.  Due to this, it is possible that the labels used to train new models are incorrect, and this easily leads to  learning wrong things and concept drift \cite{siirtola2018ESANN}. In this study, The concept drift is caused by changes in the observed data, and in addition, because of the limitations of the initial training data and machine learning model performance. Due to these, the structure of data used to update the recognition models in not similar to initial training data causing mislabelled data and inaccurate drifting models.  

When manual labelling is used, similar problems do not exist. In \cite{amrani2021personalized}, \cite{mannini2018},  \cite{siirtola2019sensors}, \cite{siirtola2021context} and  \cite{vakili2021incremental} it was shown that models benefit from user inputs and with them concept drift can mostly be avoided. For instance, in \cite{amrani2021personalized} and \cite{siirtola2019sensors}, human intelligence was used to label instances in such cases where posterior value of the prediction was low. It was noted that already by replacing a small number of uncertain labels with true labels, the accuracy of the online learning model can be improved significantly. However, as the data labelling is always burdensome, user should be bothered as seldom as possible and need for user inputs should be minimized. Therefore, as user inputs are needed in the model personalizing process to avoid concept drift, it needs to be studied is what is the best and the most effective way for human AI collaboration in the labelling process, and when user inputs really are needed and when predicted labels can be trusted.

\begin{figure*}[h]
\begin{center}
\includegraphics[width=12cm]{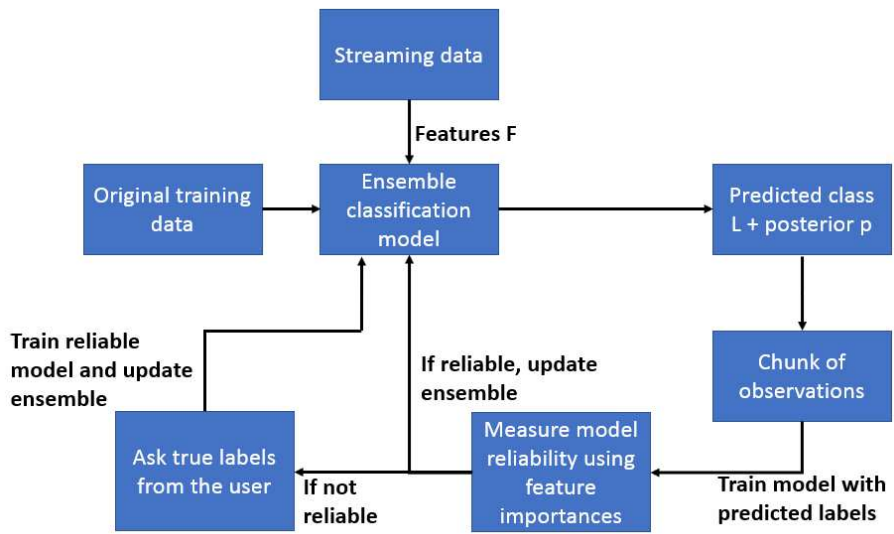}
\caption{A method to personalize human activity recognition model and reducing the need for user inputs by measuring base model reliability using feature relevance analysis.}
\label{processNew}
\end{center}
\end{figure*}

In this article, concept drift detection is based on feature relevance, a.k.a. feature importance, analysis which is not a new idea \cite{henke2015analysis} but in this article it is used for the first time in the context of human activity recognition. When it comes to explaining the reason for concept drift, there are articles where the type of drift is detected, for instance if the detected drift was virtual or real concept drift \cite{kulkarni2014incremental}, but not many studies to explain the actual reason for the drift. However, in \cite{263854} the concept drift related security applications were detected, and the reason drifting samples were explained by identifying a feature set having values that are different to affected samples and normal samples, and using this information to find the reason for the drift.

In this article, it is studied how feature relevance analysis is not only used to detect when drift is happening, but also to detect the actual reason for the drift when a limited number of possible reasons for the concept drift are predefined. With the help of this information, human-AI collaboration can be made more effective as AI can explain what was wrong with the training data and suggest which instances human needs to label manually. Human activity recognition is used as a case study.

The article is organized as follows: The data used in the experiments is explained in Section \ref{data}, Section \ref{aim} introduces the idea of the proposed method, Section \ref{case} explains the experimental setup, and results and discussion. Finally, conclusions and future work are in Section \ref{conclusion}.

\section{Experimental dataset}
\label{data}

This study uses open data set presented in \cite{Shoaib} in the experiments. It contains inertial sensor data (accelerometer, magnetometer and gyroscope, sampling rate 50Hz) from 10 persons, seven daily activities (walking, sitting, standing, jogging, biking, walking upstairs and downstairs), and five body locations. However, this study uses only accelerometer and gyroscope data from one position (arm).

Features for the study were extracted from windows of size 4.2 second (1.4 second slide), signals where these windows were calculated were raw accelerometer and gyroscope signals, as well as from magnitude signals and signals where two out of three accelerometer and gyroscope signals were square summed. Features extracted from windows include standard deviation, minimum, maximum, median, and different percentiles (10, 25, 75, and 90). Moreover, the sum of values above or below a given percentile (10, 25, 75, and 90), square sum of values above or below these percentiles, and number of crossings above or below these percentiles were used as features as well as features from the frequency domain. Altogether 131 features were extracted. From these, SFS (sequential forward selection) was used to select twenty most important ones for user-independent model, and these twenty were also used to train personal models as well as in feature relevance analysis.

\section{Feature relevance analysis to explain concept drift}
\label{aim}

This section is divided into two parts. Subsection \ref{personalizing} explains how Learn++ -based incremental learning method can be used to personalize human activity recognition models, and Subsection \ref{detect} introduces how feature important analysis can be used to detect and explain unreliable base models.

\subsection{Personalizing human activity recognition model using incremental learning}
\label{personalizing}

A incremental learning-based semi-supervised personalizing method for human activity recognition models was introduced in \cite{siirtola2019sensors}. As an incremental learning algorithm this method uses Learn++ \cite{polikar2001learn++} which is an ensemble method processing the incoming streaming data as chunks, and for each chunk a new group of weak base models are trained. These are then added to a group of previously trained base models and this combination is used as an ensemble model \cite{hammerchoosing}. The method contained three phases. Phase 1 is to train a user-independent recognition model based on data gathered beforehand. This is used as the first base model of Learn++, and at first the incoming streaming data is classified based on this. In Phase 2, the personalizing of the model starts by extracting features from the first chunk of user's personal streaming data. This data is classified using the ensemble model, and used to train new base models. If based on the posterior value the prediction provided by ensemble model is reliable, these predicted labels are used in the model training process. However, in order to avoid concept drift caused by wrong labels, training samples are hand labelled by the user if posterior shows that the prediction is not reliable. After this, new personal base model is trained and added to the ensemble. Phase 3 is similar Phase 2 but with a new data chunk. According to the results provided in \cite{siirtola2019sensors}, this method improves the recognition rates and depending on the used base classifier, user needs to hand label about 10\% of the observations.

Therefore, the reason why user inputs are needed is concept drift which means that the performance of the recognition model starts to drop. In this case, concept drift happens if wrong labels are used in the model training process of new base models, the trained model is inaccurate and eventually this leads to a situation where the ensemble model does not work at all. In \cite{siirtola2019sensors} falsely classified observations were detected based on low posterior values, and user had to label these manually. However, low posterior value does not always mean that the predicted label is incorrect. Due to this, user needs to label also observations which are correctly labelled by the ensemble method. In this article, it is studied how feature relevance analysis could help in detecting unreliable personal base models instead of detecting unreliable observations. In addition, by recognizing the reason for unreliable base model, the need for user labelled instances can be reduced even more. The method for detecting and explaining concept drift is introduced in the next subsection.

\subsection{Method to detect and explain concept drift}
\label{detect}

The proposed approach to use feature relevance analysis to understand base model reliability is shown in Figure \ref{processNew}. It is similar to approach presented in \cite{siirtola2019sensors} with some modifications. In Phase 2, posteriors are not studied in this case. Instead, new model is trained based on the obtained training data and predicted labels, and after this feature relevance analysis is used to analyze the quality of the trained personal base model based on how relevant different features are for this base model. Then if it is noted that base model is not reliable, the reason for unreliability can be explained based on feature relevance analysis as well when typical reasons for concept drift are known, and true labels for needed observations are asked from the user.

Before it is possible to detect and explain concept drift, the following steps need to be taken:
\begin{enumerate}
    \item Train a clean model using true labels.
    \item Define the most typical reasons for concept drift, the one's the aim is to detect automatically. 
    \item Train worst-case scenario model, a drifting model, for each concept drift reason defined in Step 2.
    \item Compare feature relevance of the models trained in Steps 1 \& 3.
    \item Identify a set of features for each studied concept drift reason having an unique effect to feature relevance.
\end{enumerate}

The quality of a new model can be studied in the following way: (1) if the feature relevance values of the studied model are similar to the clean model, for each identified set of features, the studied model is reliable, and (2) if the similarity is low, the studied model is suffering from the concept drift. In addition, as each studied reason for concept drift has a unique effect to the feature relevance, it is possible to explain the reason for the detected concept drift.

\begin{table*}[h]

\caption{Feature relevance values for the clean model, and the relevance differences of worst-case scenario models for S1, S2 and S3 compared to the clean model. Feature relevance differences used to detect S1 and S2 are shown in bold.}

\begin{center}

\begin{tabular}{lcccccccccc}
    Scenario & \multicolumn{10}{c}{Feature relevance differences to clean model}   \\
    \hline
    & F1 & F2&F3&F4&F5&F6&F7&F8&F9&F10\\
    \multirow{4}{*}{Clean model}  & 0.14   & 0.04   & 0.10  &  0.02   & 0.05   & 0.01 &   0.04  &  0.06  &  0.04  &  0.003  \\
    &F11&F12&F13&F14&F15&F16&F17&F18&F19&F20 \\
    & 0.07   & 0.1  &  0.01  &  0.02  &  0.02 & 0.08  &  0.03  &  0.01 &    0.01  &  0.02 \\
    \hline
    & F1 & F2&F3&F4&F5&F6&F7&F8&F9&F10\\
    \multirow{4}{*}{$diff(S1, clean)$} & 0.97  &  0.89  &  0.96  &  0.78  &  0.93  &  0.35  &  0.89  &  0.94 &   0.89  &  0.45  \\
    &F11&F12&F13&F14&F15&F16&F17&F18&F19&F20 \\
    &  0.93  &  0.96  &  0.69  &  0.86  &  0.79&  0.95 &   0.87 &   0.50  &  0.56  &  0.80 \\
    \hline
    & F1 & F2&F3&F4&F5&F6&F7&F8&F9&F10\\
    \multirow{4}{*}{$diff(S2, clean)$} & 0.21 &  \textbf{-0.66} &   0.34  & \textbf{-1.39 } & -0.46 &  \textbf{-3.32 } &  0.50 &  -0.28 &   -1.08  & -0.99   \\
    &F11&F12&F13&F14&F15&F16&F17&F18&F19&F20 \\
    & -0.50  &  0.36 &  -0.69  & -0.32  & -0.14 & 0.47  &  0.21 &  -1.71  & -0.41  &  0.08\\
    \hline
    & F1 & F2&F3&F4&F5&F6&F7&F8&F9&F10\\
    \multirow{4}{*}{$diff(S3, clean)$}  & -0.10  &  0.29  & -0.13 &   0.22  & -0.25  &  0.83 &  -0.04 &  -0.41  & -0.99&   -0.25 \\
    &F11&F12&F13&F14&F15&F16&F17&F18&F19&F20 \\
    &  -0.39 &  \textbf{-0.06 } & -0.32  & -0.98 &  -0.11 & 0.02  & \textbf{-0.20}  & -0.77  & -0.27  & \textbf{-0.30}\\
\end{tabular}
\label{weights}

\end{center}
\end{table*}

\begin{table}

\caption{Features selected to detect concept drift in each scenario, and how the sum of feature relevance differences behaves in different scenarios.}

\begin{center}

\begin{tabular}{lcccc}
    Scenario & Features &S1 & S2 & S3  \\
    \hline
    S1 & All & $\uparrow$ & $\downarrow$ & $\downarrow$\\
    S2 & 2, 4, 6 & $\uparrow$ & $\downarrow $ &$\uparrow$\\
    S3 & 12, 17, 20 & $\uparrow $ & $\uparrow$ &$\downarrow$\\
\end{tabular}
\label{rules}

\end{center}
\end{table}

\begin{figure*}[h]

\begin{center}
        \subfigure[S1]{\label{s1}\includegraphics[width=5.7cm]{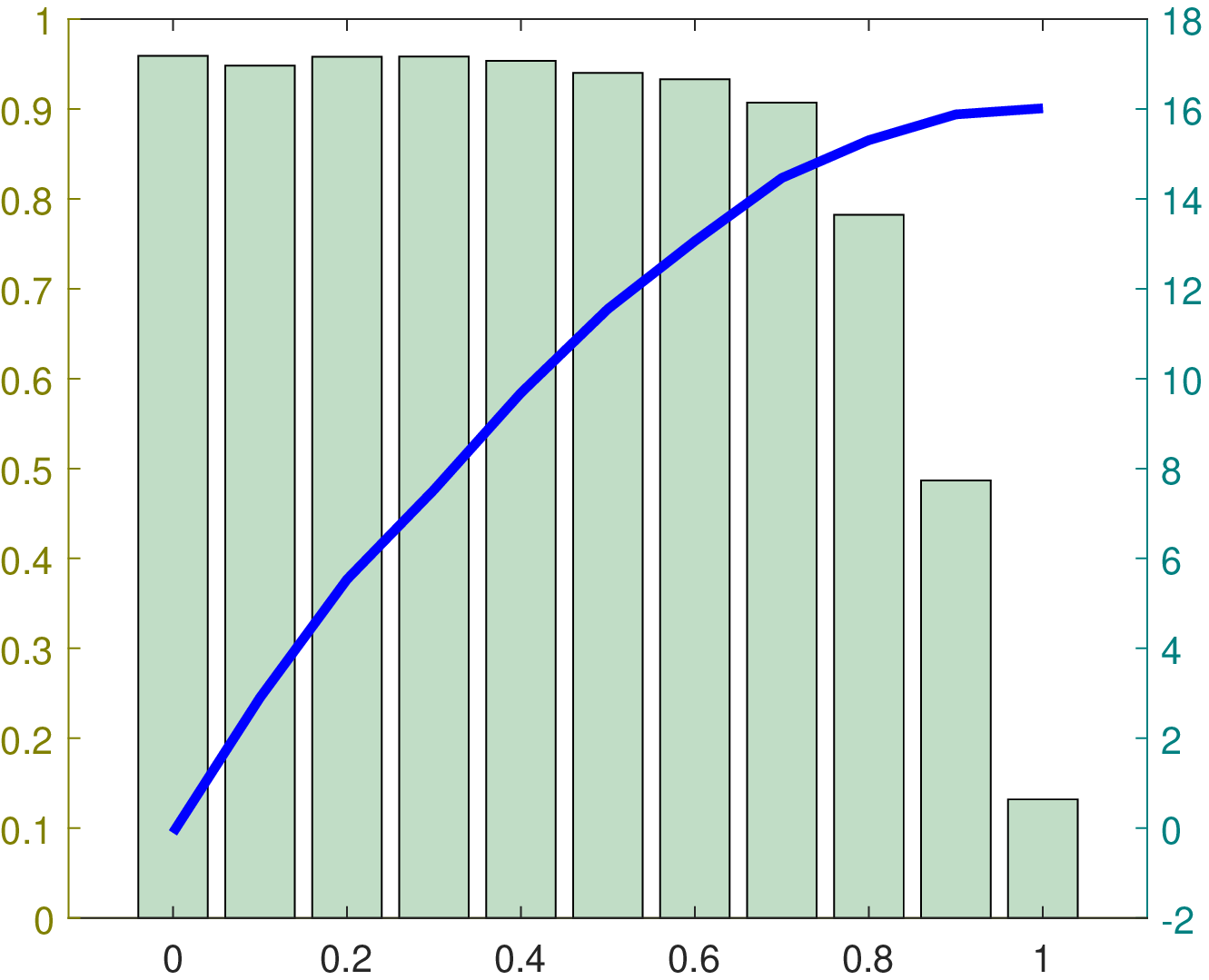}}
		\subfigure[S2]{\label{s2c1}\includegraphics[width=5.7cm]{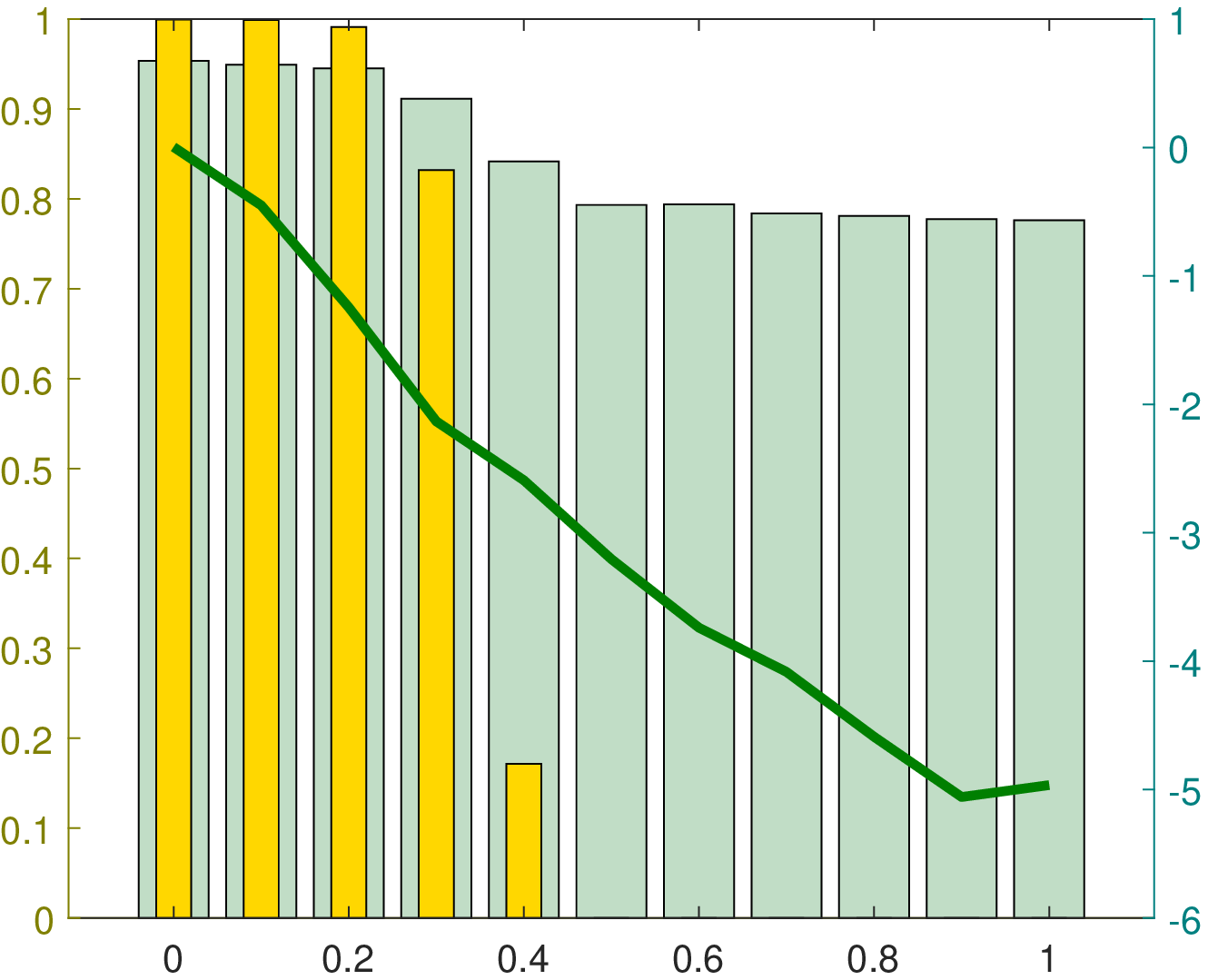}}
		\subfigure[S3]{\label{s2c2}\includegraphics[width=5.7cm]{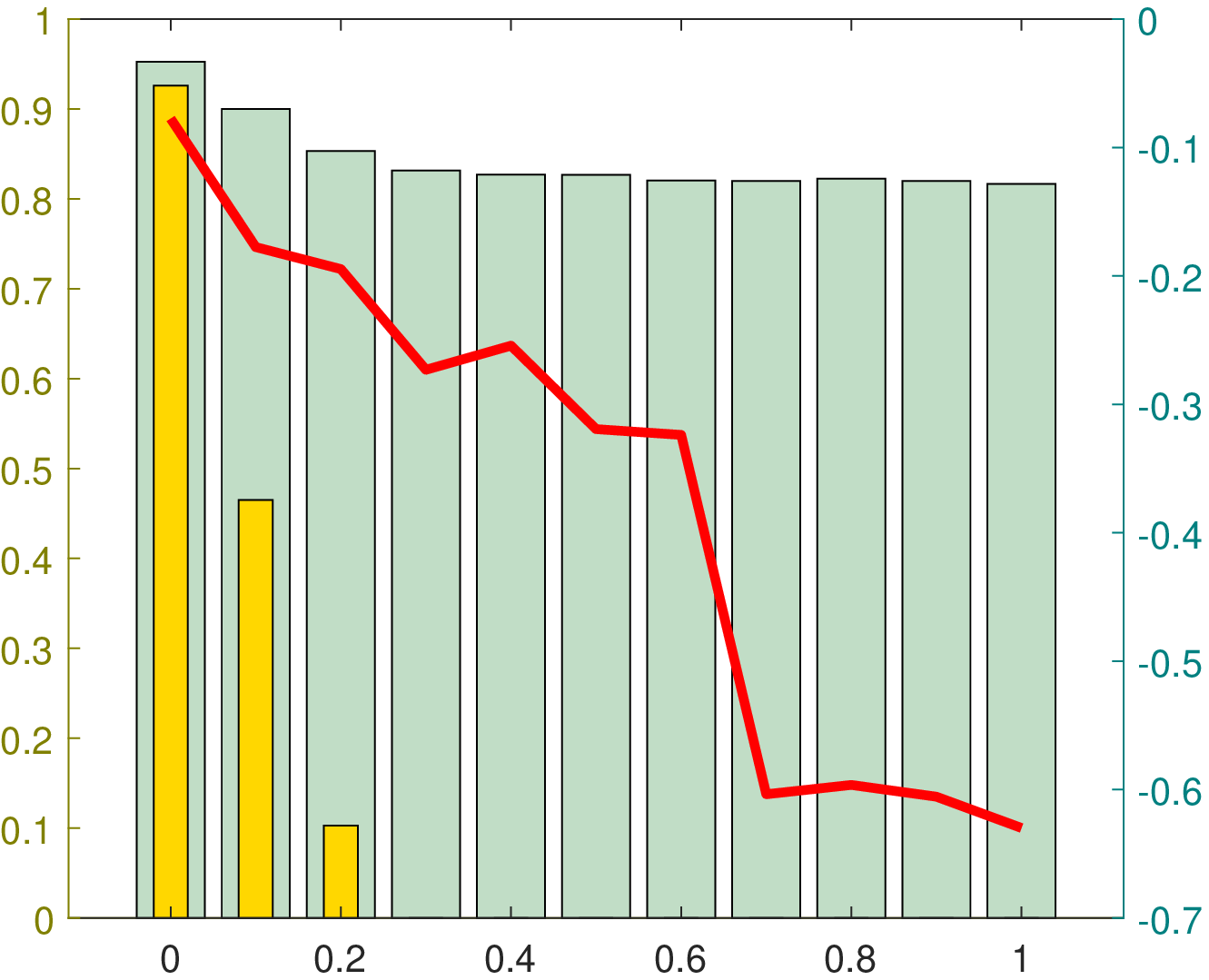}}

    \caption{The effect of concept drift in studied scenarios. x-axis shows how big ratio of the labels are randomly labeled. Solid line shows the sum of relevance differences between the studied model and the  clean model. Green bars show how the studied model performs in the test data. In S2, yellow bars show the recognition rate of walking-activity in test data, and in S3 yellow bars show the recognition rate of walking downstairs-activity in test data.} 

\label{results}
\end{center}
\end{figure*}

\section{Case study: Human activity recognition data to detect and explain concept drift}
\label{case}

Experimental setup used in this article is explained in Subsection \ref{exper}, and the results obtained using this as well as the discussion are in Subsection \ref{res}.

\subsection{Experimental setup}
\label{exper}

For the experiments, each person's data are divided into three parts so that each part contains the same amount of observations from each activity \cite{siirtola2018ESANN}. To train personal base models, two first parts were combined and used for training, and the last one was only used for testing.

In the first part of the experiment, a clean personal model is trained using almost perfect labels containing 95\% of true labels and 5\% of randomly labeled data (Step 1). Secondly, three reasons for the concept drift that are studied in this article are pre-defined (Step 2): (S1) each class of the training data contains approximately equal amount of false labels, (S2-S3) one class of the training data is mixed with another class so one class of the training data contains a lot of false labels: (S2) the labels of walking are mixed with biking, and (S3) the labels of walking upstairs are mixed with walking downstairs. After this, a worst-case scenario model is trained for each scenario S1-S3 (Step 3): worst case scenario for S1, all the samples used in the training are randomly labelled;  for S2, all the training samples of biking are mislabeled as walking; and for S3, all the training samples of walking upstairs are mislabeled as walking downstairs. Experiment were performed using Matlab 2018b, and all the models were trained using Bootstrap-aggregated classifier.

To compare feature relevance of clean and worst-case models (Step 4), feature relevance for the models were obtained using Matlab's \textit{predictorImportance}-function, and Table \ref{weights} shows the average relevance of each feature of clean model when personal model was trained for each subject. In addition, Table \ref{weights} shows how many percentages the feature relevance of three worst-case scenario models differ from the clean model. These differences are calculated using Equation $diff(S, clean) = \frac{F_i^{clean} - F_i^{S}}{F_i^{clean}}$
where $F_i^{clean}$ is the relevance of the $i$th feature of the clean model $clean$, and $F_i^{S}$ is the relevance of the $i$th feature of the other model $S$, in this case the worst-case scenario model. Idea is to find  unique feature relevance combinations based on the Table \ref{weights}, so that the feature relevance sum can be used to identify the drift and the cause for this. When signs of the relevance differences are studied, it can be noted for instance that the relevance difference of feature number 2 (F2 in the Table \ref{weights}) is negative only in scenario S2 and positive for S1 \& S3. 
Based on searching this type of unique feature behaviours, Table \ref{rules} shows which feature combinations are manually selected to detect and explain drift in each scenario (Step 5). For S1 relevance sum is positive when all the features are selected, for scenarios S2 \& S3 this sum is negative. The sum of features 2, 4 and 6 is only negative in S2, and the sum of features 12, 17 and 20 is only negative in  S3. 

\begin{figure*}[h]

\begin{center}
        \subfigure[S1]{\label{s1vs}\includegraphics[width=5.7cm]{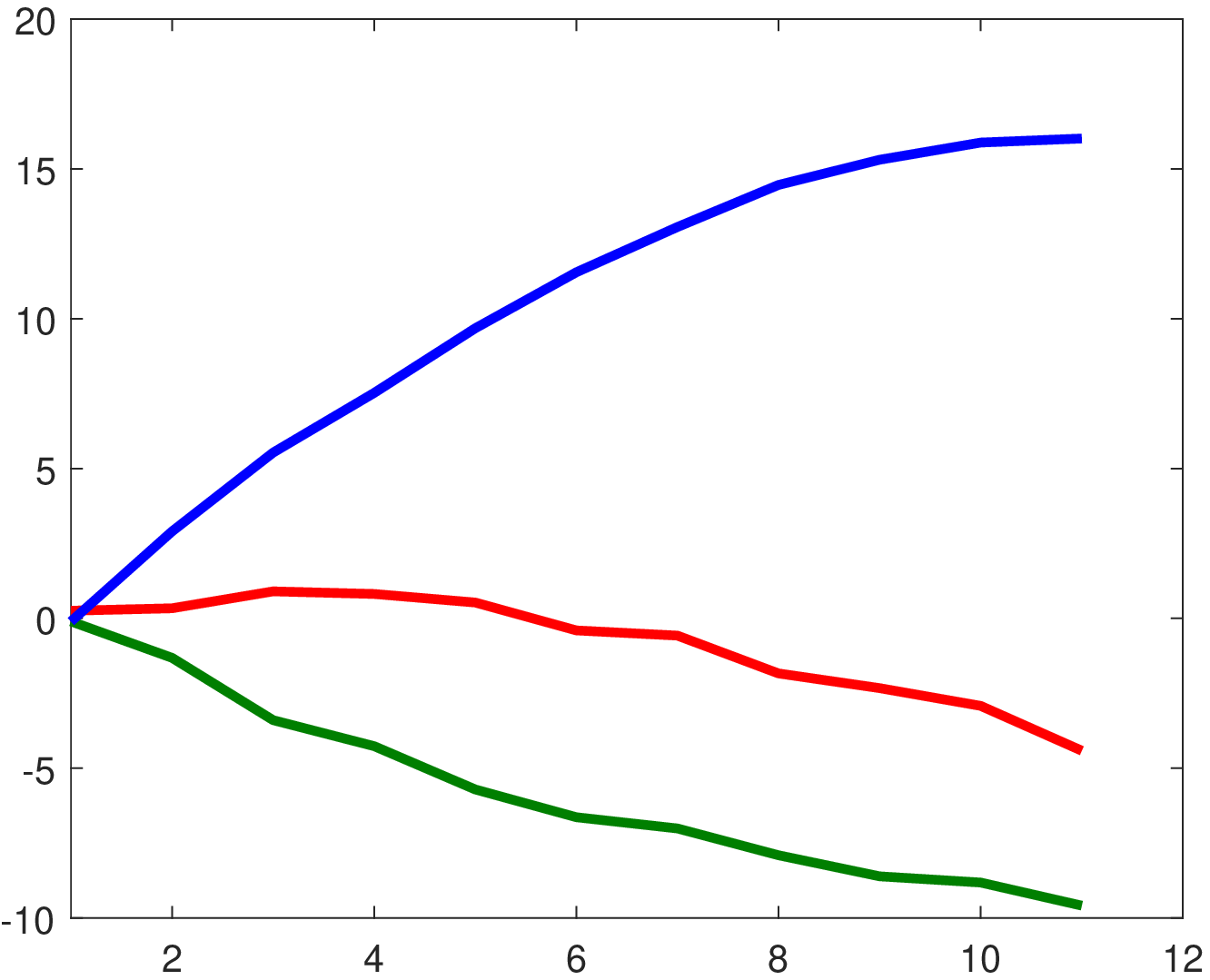}}
		\subfigure[S2]{\label{s2c1vs}\includegraphics[width=5.7cm]{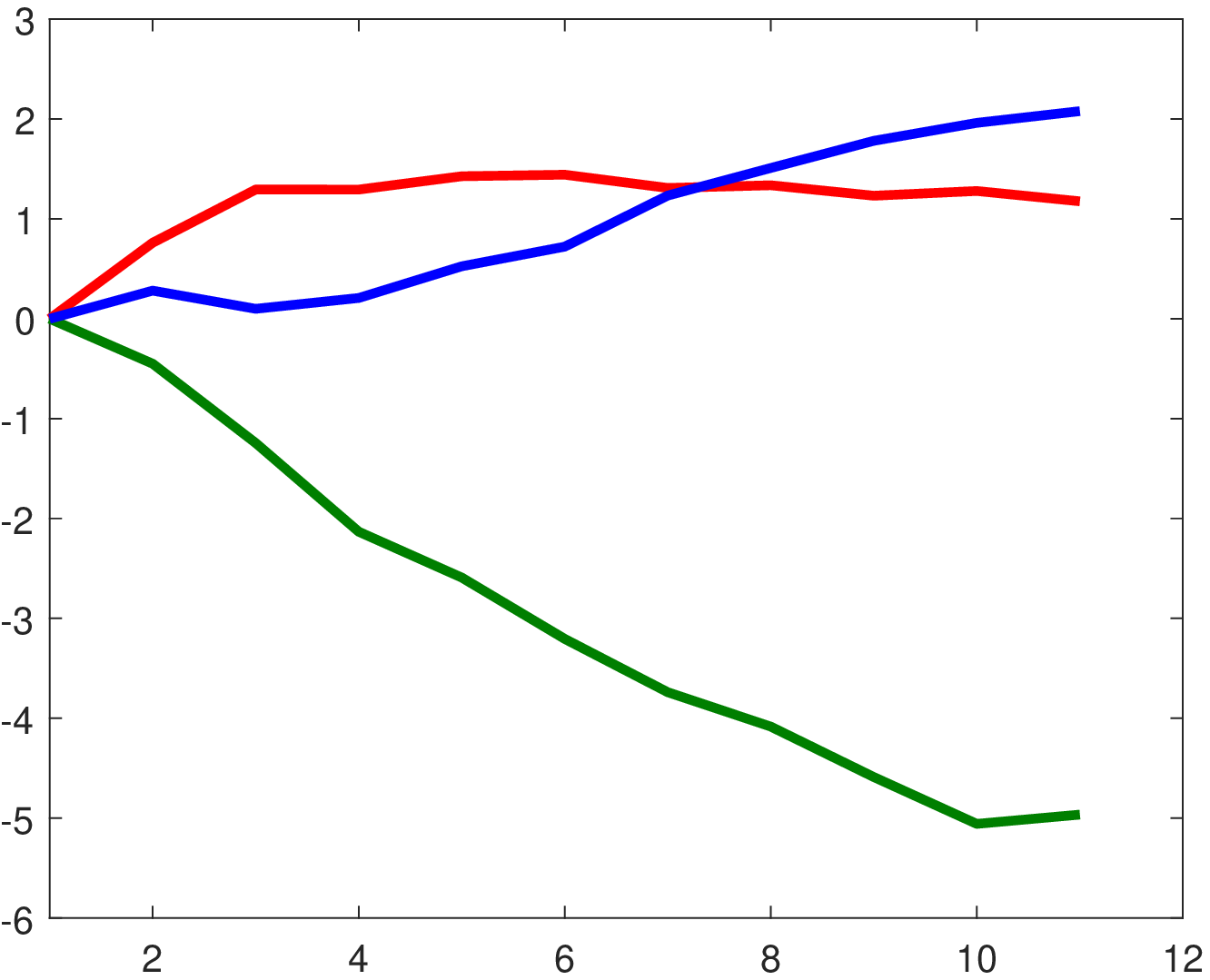}}
		\subfigure[S3]{\label{s2c2vs}\includegraphics[width=5.7cm]{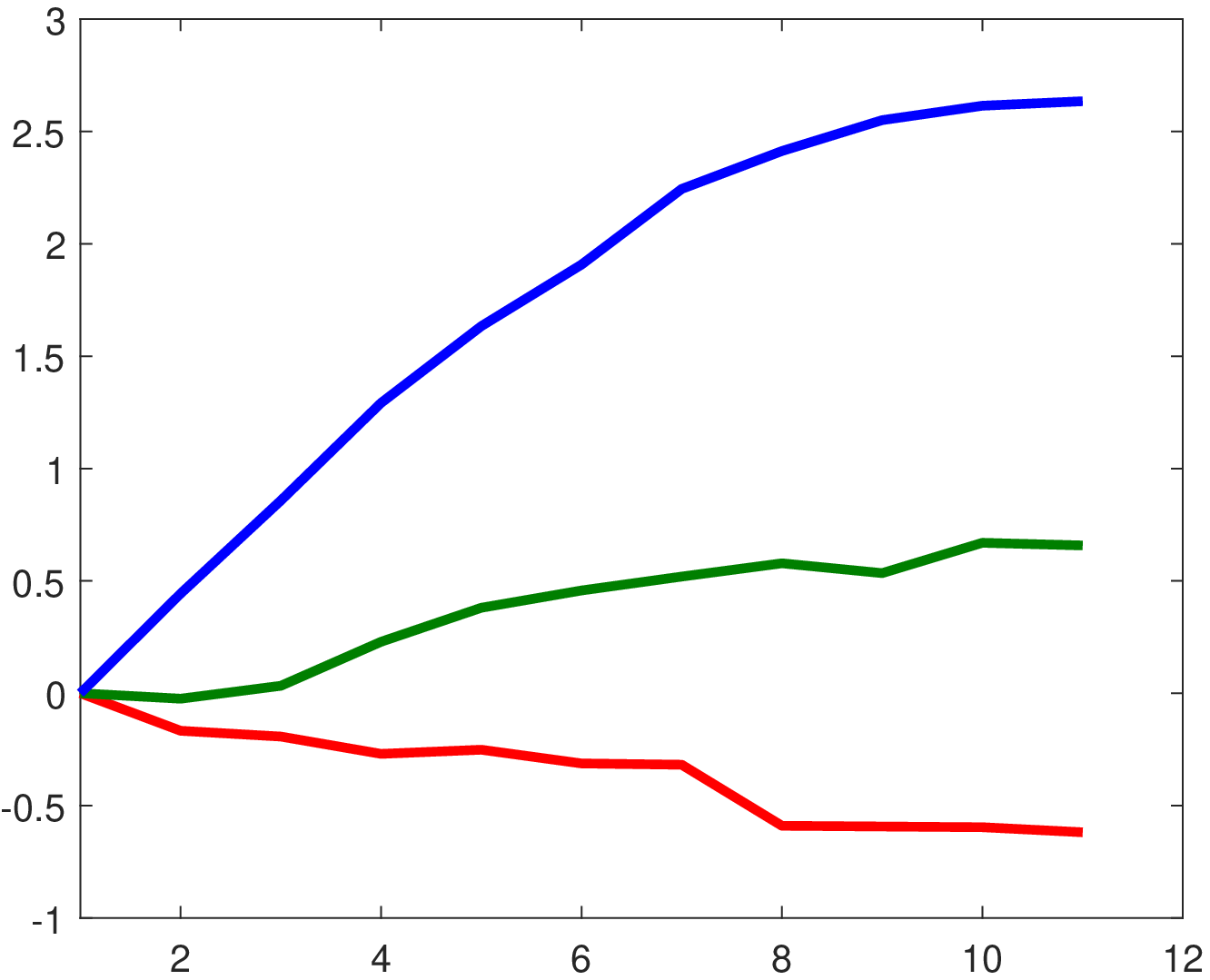}}

    \caption{Explaining concept drift. Each studied reason for the concept drift has a unique effect to the relevance of the selected features, and features selected to detect concept drift in certain scenario behaves differently compared to relevance sum of the features selected for other two scenarios. x-axis shows how big ratio of the labels are randomly labeled. Blue line represents the sum of relevance differences selected for S1, green line represents the same for S2, and red line for S3.} 

\label{compare}
\end{center}
\end{figure*}

\subsection{Results and discussion}
\label{res}


The results of Figure \ref{results} show how the relevance sum of the selected features (see Table \ref{rules}) for each scenario behave when the data quality used in the training process of base model reduces more and more, and therefore, the base model performance starts to gradually drift. In Figure \ref{s1}, the x-axis shows how big ratio of the labels used in the training process are randomly labeled, in Figures \ref{s2c1} and \ref{s2c2} it shows how big ratio of the training data is mixed with another class. The results from scenario S1 (Figure \ref{s1}) show that if the recognition rate of the model used to predict labels for the training data drops equally for each class, and therefore the training data of the studied model has approximately equal amount of false labels for each class, it has surprisingly small impact on the the model performance on test set (green bars). Still, when the distance between feature relevance differences obtained from a model trained using correct labels and a model trained using false labels is calculated, there is a clear difference in the relevance sum of the selected features when the number falsely labeled training samples increases (blue line). However, when it comes to human activity recognition, scenarios S2 and S3, where model performance drops only for one class, is more typical in real-life scenarios. Figure \ref{s2c1} shows how based on feature relevance analysis it is possible to detect situation where the labels of biking and walking are mixed in the training data, and Figure \ref{s2c2} shows results when the labels of walking upstairs are mixed with walking downstairs in the training data. In both cases, it can be noted that the more labels are mixed, the bigger effect it has to the relevance sum of the selected features (green line in Figure \ref{s2c1}, and red line in Figure \ref{s2c2}). The sum of relevance differences is smaller in S3 than in S2 (see right y-axis in Figures \ref{s2c1} and \ref{s2c2}), but  also in the case of S2 false labels has a clear effect to  the relevance sum of the selected features when more and more false labels are used in the model training process. However, still this smaller difference can cause problems to recognize type S3 drift. Moreover, it can be noted that in the case of S3, the mixing of walking upstairs and downstairs labels has a rapid effect on the recognition rate of walking upstairs-activity. Already, 10 percentage of false labels in the training set leads to situation where only 50\% of the walking upstairs-activity if correctly classified in the test set (yellow bars).  In the case of S2, the mixing of walking and biking-labels does not have as sudden effect on the test set performance, but still it can be seen that if 40\% of the training samples of biking-activity are false labeled, then the performance in test set is only around 20\%. Therefore, in can be noted that in each scenario, feature important analysis can detect if the training data is not of high quality when compared feature important of a model that is known to be accurate.

The results of Figure \ref{results} show that concept drift has a clear effect on sum of feature relevance difference values in each scenario. However, the main aim of this article was to study if it is not only possible to detect concept drift, but also find the reason for the concept drift. This is studied in Figure \ref{compare} where it is compared how the relevance sum of the features selected to detect concept drift in certain scenario behaves compared to relevance sum of the features selected for other two scenarios. In fact, Figure \ref{results} shows that these sums behave in each scenario as assumed based on Table \ref{rules}. In the Figure \ref{s1vs}, as assumed based on Table \ref{rules}, only the sum of features selected for S1 is positive and goes up (blue line) (and the sum of features selected for S2 \& S3 goes down) when the concept drift of scenario S1 is affecting to training label quality. Similarly, in the Figure \ref{s2c1vs} only the sum of features selected for S2 is negative (green line), and the sum of features selected for S1 \& S3 is positive, when concept drift is caused by mixing of walking and biking as it happens in scenario S2. Moreover, the Figure \ref{s2c2vs} shows that only the sum of features selected for S3 is negative (red line), when mixing of upstairs and downstairs is the reason for concept drift (scenario S3). This means that by following the rules given in Table \ref{rules}, the feature relevance analysis can be used to explain the reason for the concept drift when a limited number of possible reasons for concept drift are predefined. This helps human to understand when there is a need to label observations by hand. For instance if it is known that the reason for concept drift is the mixing of walking and biking-activities, human does not need to label the whole training data chunk but only observations related to walking and biking-activities.

\section{Conclusions and future work}
\label{conclusion}

Human activity recognition model can be personalized based on Learn++ by learning from streaming data and adding personal base models to ensemble. The problem is that in order to avoid concept drift caused by learning wrong things, user inputs are needed. This article presented a feature relevance analysis-based method to detect unreliable base models to avoid concept drift and to reduce the need for user inputs. Moreover, as a main results of this article, it was shown that the reason for concept drift can be explained when a limited number of possible reasons for the concept drift are predefined. This helps human to understand when there is a need to label observations by hand, as for instance if it is known that the reason for concept drift is the mixing of walking and biking-activities, human does not need to label the whole training set but only observations related to walking and biking-activities. However, the presented method is only a one piece in the puzzle when it comes to solving this labeling problem, in fact, in order to minimize the need for human inputs the combination of the presented methods, posterior analysis \cite{siirtola2019sensors} and data distribution analysis \cite{concept_drift} should be studied.


This study has limitations. The experiments were performed using a small dataset, and it was not possible to study how the natural variance effects to the feature relevance. In addition, it should be studied how to define the rules presented in Table \ref{rules} automatically.  Moreover, different classifiers give different relevance for the features \cite{saarela2021comparison}, therefore, the presented approach does not necessarily work with every classifier. However, small experiments were performed using AdaBoost-algorithm, and these were supporting the findings of this article. In addition, more powerful similarity measures should be studied as well as how to select features not just to maximize the recognition rates, but also to maximize concept drift detection


\bibliographystyle{ACM-Reference-Format}
\bibliography{Template}


\end{document}